\documentclass{article}

\usepackage{algorithm}
\usepackage{nicefrac, xfrac}
\usepackage{amsmath,amssymb}
\usepackage{graphicx}
\usepackage{tabularx}
\usepackage{subcaption}

\DeclareMathOperator{\EX}{\mathbb{E}}

\usepackage[numbers]{natbib}
\usepackage[preprint]{neurips_2023}

\usepackage[utf8]{inputenc} 
\usepackage[T1]{fontenc}    
\usepackage{hyperref}       
\usepackage{url}            
\usepackage{booktabs}       
\usepackage{amsfonts}       
\usepackage{nicefrac}       
\usepackage{microtype}      
\usepackage{xcolor}         

\title{Learning Discrete Weights and Activations Using the Local Reparameterization Trick}

\author{%
  Guy Berger \\
  Bar-Ilan University, Israel\\
  \texttt{guy.berger1@biu.ac.il} \\
  \And
  Aviv Navon \\
  Bar-Ilan University, Israel\\
  \texttt{aviv.navon@biu.ac.il} \\
  \AND
  Ethan Fetaya \\
  Bar-Ilan University, Israel\\
  \texttt{ethan.fetaya@biu.ac.il} \\
}

\begin{document}

\maketitle

\begin{abstract}
In computer vision and machine learning, a crucial challenge is to lower the computation and memory demands for neural network inference. A commonplace solution to address this challenge is through the use of binarization. By binarizing the network weights and activations, one can significantly reduce computational complexity by substituting the computationally expensive floating operations with faster bitwise operations. This leads to a more efficient neural network inference that can be deployed on low-resource devices. In this work, we extend previous approaches that trained networks with discrete weights using the local reparameterization trick to also allow for discrete activations. The original approach optimized a distribution over the discrete weights and uses the central limit theorem to approximate the pre-activation with a continuous Gaussian distribution. Here we show that the probabilistic modeling can also allow effective training of networks with discrete activation as well. This further reduces runtime and memory footprint at inference time with state-of-the-art results for networks with binary activations.


\end{abstract}

\section{Introduction}
\label{sec:intro}

 As neural networks become more powerful and their applications more commonplace, there is an important and growing need to reduce computational costs and memory requirements. This is especially important in edge devices, e.g., smartphones, that have weaker processors and need to optimize energy consumption. To address this challenge, one promising approach is to train binary or ternary networks, where the weights are constrained to take on only a small number of discrete values~\citep{BinaryConnect, TWN, TTQ, SYQ, Fixed}. One can also binarize the activations (e.g., using the sign function) to further improve the efficiency of the network, albeit with a larger reduction in accuracy~\citep{XNOR, BNN, DoReFa, ABC, BirealNet, lottery, RAD, IRNet, AutoBNN, DSQ}.\\

Most previous works that have utilized sign activation directly apply it during the forward pass and use heuristic or approximation methods to estimate the gradients during the backward pass, e.g., straight-through gradient or other biased gradient estimators \citep{BNN, BinaryConnect, XNOR, IRNet, BNNplus}.
In this paper, we propose a novel method to compute gradients for networks with discrete activation by employing a smooth approximation. We construct a fully differentiable probabilistic model to approximate the discrete network during training. After training, we sample from our trained probabilistic model to get our discrete weights.\\

Our method is based on the observation by \citet{Kingma02} which states that if the weights 
of a certain layer are sampled from independent Gaussian distributions, then one can get better stochastic gradient estimation by modeling and sampling the Gaussian distribution of the pre-activations 
 instead of weights. They called it the local reparameterization trick. 
\citet{Shayer01} extended the local reparameterization trick to train a network with discrete weights. When the weights 
are discrete and stochastic, the pre-activations 
can still be well approximated by a Gaussian distribution according to the (Lyapunov) central limit theorem (CLT). Then, by using the reparameterization trick, we can compute the derivatives of a smooth distribution.
In this work, we extend \citep{Shayer01} and train a network with discrete weights and activations. 
Based on the observation that the pre-activations 
are well approximated by a Gaussian distribution, we 
propose sampling discrete values from the induced distribution given our Gaussian approximation. We construct a probabilistic model which is fully differentiable, using the Gaussian CDF function to calculate the activation probabilities and Gumbel-Softmax \citep{Maddison01,Jang01} as a smooth approximation for categorical sampling.

We demonstrate the effectiveness of our approach on several benchmark datasets. The experiments show that our proposed method, which we name LAR-nets (Local Activation Reparameterization  networks), outperforms previous SoTA binarization approaches. To summarize, we make the following contributions: 
\begin{itemize}
    \item  Introduce a novel approach for learning with discrete activations based on the local reparameterization trick.
    \item Propose a new variant for batch normalization to extend the applicability of the normalization layer  to distribution over weights.
    \item  Demonstrate the effectiveness of the proposed approach against multiple baselines for training discrete networks on several vision benchmarks.

\end{itemize}

\begin{figure}
  \begin{subfigure}{0.37\textwidth}
    \includegraphics[width=\linewidth]{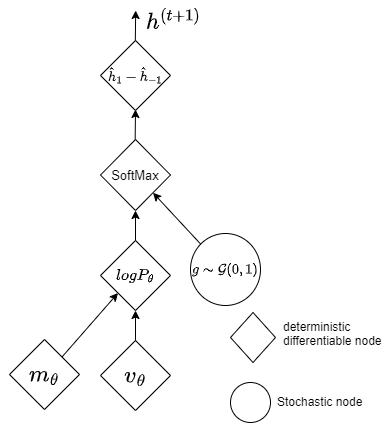}
    \caption{} \label{fig:1a}
  \end{subfigure}%
  \hspace*{\fill}   
  \begin{subfigure}{0.33\textwidth}
    \includegraphics[width=\linewidth]{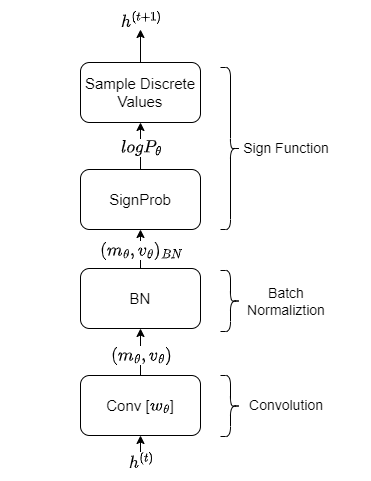}
    \caption{} \label{fig:1b}
  \end{subfigure}%
  \hspace*{\fill}   
  \begin{subfigure}{0.22\textwidth}
    \includegraphics[width=\linewidth]{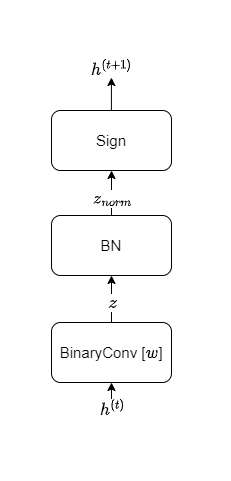}
    \caption{} \label{fig:1c}
  \end{subfigure}
\caption{
(a) The structure of our LAR-Net layer with discrete weights and activation. The reparameterization allows the gradients to flow to $\theta$ in the backward pass through all the differentiable nodes; 
(b) The structure of a basic Conv-BN-Act block during the training phase; 
(c) The structure of basic Conv-BN-Act layer on evaluation phase (after sampling weights from $\theta$); 
\label{fig:networks}
}
\end{figure}

\section{Related Work}

Binary networks have been an active area of research in deep learning over the past few years. \citet{BinaryConnect} introduced BinaryConnect, a method for training binary neural networks using binary stochastic weights. They sample binary weights and compute gradients as if it was a deterministic full-precision network.
BNN \citep{BNN} and XNOR-Net \citep{XNOR} suggested a different approach; They discretize the weights and activations during the forward pass and back-propagate through this non-continuous discretization using the \emph{straight-through} estimator.
 XNOR-Net also  introduces a continuous scaling factor to the binarized weights, which became popular in later works. Other similar works based on the straight-through estimator extend it to ternary networks (TWN \citep{TWN}, TTQ \citep{TTQ}).\\


Motivated by the high potential to reduce model complexity and improve computational efficiency by also discretizing activations, works that discretize both weights and activations became more common.
DoReFa-Net \cite{DoReFa} extended XNOR-Net to accelerate the training process using quantized gradients. ABC-Net \cite{ABC} suggested reducing the quantization error by linearly combining multiple binary weight matrices and scaling factors to fit the full-precision weights and activations. Other works suggested making modifications in the network architectures, such as adding tailored-made layers, to make them more fit to binary neural network training \cite{AutoBNN, BirealNet}.
Some recent works tried to achieve good quantization functions in forward propagation, which can reduce the gradient error as well. \citet{DSQ} presented a differential soft quantization (DSQ) that has more accurate gradients in the backward pass. 
Diffenderfer and Kailkhura \citep{MPT} proposed what they called Multi Prize Ticket (MPT). 
They incorporated a binarization scheme with weight pruning (which results in a ternary network).
For the weights and activations binarization they suggested a modified gradient estimator for the sign function. For the weights pruning, inspired by Frankle and Carbin \citep{lottery}, they proposed a scheme to prune a randomly initialized network using a learnable mask by updating pruning scores during the training. \citet{RAD}, suggested reshaping the data distribution before binarization by adding a distribution loss for learning the proper binarization (BNN-DL). 
\citet{IRNet} suggested IR-Net, which aimed to minimize the information loss by maximizing the information entropy of the quantized parameters and minimizing the quantization error. Several other works discretize the activation, but to more than one bit \citep{SYQ,hu2018hashing, Fixed}.





\section{Background}
In this section, we describe the local reparametrization trick and the LR-net approach.
\paragraph{Notation.} Let $\{(x_i,y_i)\}_{i=1,...,N}$ denote our training examples. We denote by $W$ the set of all model parameters and let \(W^{(t)}\) denote the weights matrix for layer \(t\). Furthermore, let 
\(w_{ij}^{(t)}\) denote the element of \(W^{(t)}\), i.e., $[W^{(t)}]_{ij}=w^{(t)}_{ij}$.
The pre-activations are defined as \(z^{(t+1)}=W^{(t)}h^{(t)}\), where \(h^{(t)}\) is the  activations of layer \(t\) (we omit the bias term for simplicity). We let $h^{(0)}=x$ and define
\(h^{(t+1)}=\phi(z^{(t+1)})\), where \(\phi (\cdot)\) is a non-linear activation function.\\

We assume a stochastic model in which each weight \(w_{ij}^{(t)}\)  is sampled independently from a multinomial distribution \(\mathcal{W}_{ij}^{(t)}\). The objective is minimizing the expected loss w.r.t. \(\:\mathcal{W}\), where \(\mathcal{W}\) denotes the the distribution over \(W\),
\begin{equation}\label{eq:objective}
  L(\mathcal{W})=\EX_{W\sim\mathcal{W}}[\sum_{i=1}^{N}\ell(f(x_{i},W),y_{i})].  
\end{equation}
The classic approach for minimizing Eq. \ref{eq:objective} with a discrete distribution is the log-derivative trick \citep{REINFORCE}:
	\begin{equation}
	\nabla L(\mathcal{W})=\mathbb{E}_{W\sim\mathcal{W}}\left[\sum_{i=1}^{N}\ell\left(f(x_i,W),y_i\right)\nabla\log(P(W))\right].
	\end{equation}
	While this allows us to get an unbiased estimation of the gradient, it suffers from high variance, which makes optimization using this method challenging.\\
	
A popular alternative for \emph{continuous} distributions is the \emph{reparameterization trick} \citep{Kingma01}  - instead of optimizing $\mathbb{E}_{x\sim p_\varphi}[f(x)]$ for distribution parameters $\varphi$  we parametrize $x = g(\epsilon;\theta)$ where $\epsilon$ is drawn from a known fixed distribution $p(\epsilon)$ (usually Gaussian) and optimize $\mathbb{E}_{p(\epsilon)}[f(g(\epsilon,\theta))]$ for $\theta$. To estimate the gradient w.r.t $\theta$, we sample $\epsilon_1,...,\epsilon_m$ and use the Monte-Carlo approximation:
	\begin{equation}
	\nabla_\theta \mathbb{E}_{p(\epsilon)}[f(g(\epsilon,\theta))]\approx\sum_{i=1}^{m}\nabla_\theta f(g(\epsilon_i,\theta)).
	\end{equation}

\citet{Kingma02} proposed an alternative for the task of 
variational approximation for Bayesian neural networks. The authors make the key observation that if a weight matrix $W^{(t)}$ is sampled from an independent Gaussians $w_{ij}\sim\mathcal{N}(\mu_{ij},\sigma^2_{ij})$, then the pre-activations $z^{(t+1)}=W^{(t)}h^{(t)}$ are distributed according to \begin{equation}\label{eqn:local}
	z_i^{(t+1)}\sim\mathcal{N}\left(\sum_{j}\mu^{(t)}_{ij}h^{(t)}_j,\sum_j{\sigma_{ij}^{(t)}}^2 {h_{j}^{(t)}}^2\right)
\end{equation}
 
This allows sampling the pre-activations instead of the model weights, which results in lower-variance gradient estimations \citep{Kingma02} and better optimization. This approach is termed the \emph{local reparameterization trick}.\\ 

The local reparameterization trick has been further extended to  training networks with \emph{discrete} weights. 
\citet{Shayer01} introduced a method to train networks with binary \(\{\pm1\}\) or ternary \(\{-1,0,1\}\) weights (but it can be applied to a wider range of discrete values). The key idea behind this method is that while the pre-activations  \(z_i=\sum_{j}w_{ij} h_{j}\) \footnote{We omit the layer index $(t)$ from that point to simplify the notation.} are discrete, from the  (Lyapunov) central limit theorem (CLT) they are well 
approximated by the  Gaussian distribution \(z_i \sim \mathcal{N}(\sum_{j}\mu_{ij}h_j,\,\sum_{j}\sigma_{ij}^{2}h_{j}^{2})\). 
We note that  \(\mu_{ij}\) and \(\sigma_{ij}\) are now the mean and variance of a multinomial distribution, and not the mean-field Gaussian distribution. 
By sampling \(\epsilon_{i} \sim \mathcal{N}(0,1)\) we can represent the output as \(z_i=m_{i}+\epsilon_{i}\cdot v_i\),
where \(m_i=\EX(z_i)=\sum_{j}\mu_{ij}h_j\)
and \(v_{i}^{2}=Var(z_i)=\sum_{j}\sigma_{ij}^{2}h_{i}^{2}\).
Let \(\theta_{ij}\) be the parameters of the multinomial distribution over \(w_{ij}\), then, our goal is to minimize the expected loss \(\ell\) \(w.r.t.\) \(\theta\) (where $\theta$ denotes the set of all parameters $\theta_{ij}$),
\begin{equation}
  \nabla_{\theta} L(\mathcal{\theta})=\EX_{p(\epsilon)}[\sum_{i=1}^{N}\ell(f(h_{i},\epsilon,\theta),y_{i})].  
  \label{eqn:loss}
\end{equation}

\section{Our Method}
In this section, we describe our LAR-net approach. We propose a novel extension to the LR-net \citep{Shayer01} that allows for learning discrete activations. Furthermore, we describe a novel batch-normalization layer to complement our method. Finally, we outline the inference procedure using our method.

\subsection{Learning Discrete Activations}

Here we propose a novel extension to \citep{Shayer01} for training networks with discrete weights and activations. Assume we have a network with a distribution over discrete weights and the $sign$ function as its non-linearity. If we approximate the pre-activation output \(z_i\)  by the Gaussian distribution, we can directly calculate the distribution of the values of the discrete activation \(sign(z_i)\) using the probability that the sampled Gaussian would be positive


\begin{equation}
  sign(z_i)=\begin{cases}
    +1, & \text{\(with \; probabailty \:\: p_i=1-\Phi(-\frac{m_{i}}{v_{i}})\)}.\\  
    -1, & \text{\(with \; probabailty \:\: 1-p_i=\Phi(-\frac{m_{i}}{v_{i}})\)}.   
  \end{cases}
\end{equation}
where $\Phi$ is the Gaussian CDF function. We note that the distributions of $sign(z_i)$ are independent, as each one depends on a different row in the matrix $W$. We also note that if we tried to compute the distribution on subsequent layers, the outputs would not be independent anymore due to the shared inputs. As this would impair our CLT approximation, we instead choose to sample the discrete activation at each layer.   \\ 

In order to differentiate through the discrete activation sampling, we use the Gumbel-Softmax approximation \citep{Maddison01,Jang01}. This method builds on the Gumble-Max trick \citep{Gumbel}, presented in eq. \eqref{eq:gumbel}, that showed one can get a categorical sample by taking the maximum of the logits with additional Gumbel distributed noise,

\begin{equation}\label{eq:gumbel}
  \hat{h}=one\_hot(\arg \max_{k} [g_k+\log\pi_k]), \:\:\:\: g_k \sim Gumbel(0,1) 
\end{equation}

The approximation in \citep{Maddison01,Jang01} replaces  the discrete max with a smooth Softmax function with a temprature parameter $\tau$. It can be shown that as $\tau\rightarrow 0$ the Softmax converges to the max and we get a sample from the desired categorical distribution.

 
\begin{equation}
\label{eqn:gumbel_Softmax}
  \hat{h}_k=\frac{exp((log(\pi_k)+g_k)/\tau)}{\sum^{K}_{j=1}exp((log(\pi_j)+g_j)/\tau)} \:\:\:\: for \:\: k=1...K
\end{equation}

We are interested in discrete variables with two classes (\(\{\pm1\}\)). We define the class probabilities vector as \(\pi_i=[1-p_i,p_i]\). Using eq. \eqref{eqn:gumbel_Softmax} we can generate a samples vector \([\hat{h}_{i,-1},\hat{h}_{i,1}]\) (where $\hat{h}_{i,-1}+\hat{h}_{i,1}=1$) and then multiply it with our discrete values $[-1, 1]$ (each discrete value is multiplied with the corresponding element in $\hat{h}_i$), resulting in a very simple formula:

\begin{equation}
  h^{(t+1)}_i= \hat{h}^{(t+1)}_{i,1} - \hat{h}^{(t+1)}_{i,-1}\approx sign(z_i^{(t+1)})
\end{equation}


When $\tau\rightarrow 0$ and the Softmax is replaced with the argmax, we get the exact sign activation. We note that the Gumbel-Softmax has two variants: ``hard" and ``soft",  our reported results are with the ``hard" variant, which achieved slightly better results (see \cite{Maddison01} for details).\\ 

This leads to a simple algorithm for the forward pass in the training phase. Let \(\theta_{ij}\) be the 
parameters of the multinomial distribution over \(w_{ij}\). At the forward pass we compute the 
weights mean \(\mu_{ij}=\EX_{\theta_{ij}}[w_{ij}]\) and  
variance \(\sigma_{ij}^2=Var_{\theta_{ij}}[w_{ij}]\). Then, we calculate the mean and 
variance of \(z_i\), \(m_i=\sum_{j}\mu_{ij}h_j\)
and \(v_i=\sum_{j}\sigma_{ij}^{2}h_{j}^{2}\). Using the mean and variance, we calculate the discrete 
distribution of \(sign(z_i)\), and then we use the Gumbel Softmax to (approximately) sample from that distribution.
The forward pass during the training pahse is summarized in Algorithm \ref{algo:algo1}. The backward pass is straightforward, since all the functions are differentiable (see Fig. \ref{fig:1a}).

\begin{algorithm}
    \caption{Discrete Weights and activation layer}
    \textbf{Input:} \(h \in \mathbb{R}^{d_{in}} \) \\
    \textbf{Parameters:} Multinomial parameters $\theta_{ij} $ for each weight \\
    \textbf{Forward Pass:} \\
        1: Compute \(\mu_{ij}=\mathbb{E}_{\theta_{ij}}[w_{ij}]\) and \(\sigma_{ij}^2=Var_{\theta_{ij}}[w_{ij}]\) \\
        2: Compute \(m_i=\sum_{j}\mu_{ij}h_j\) and \(v_i=\sum_{j}\sigma_{ij}^{2}h_{i}^{2}\) \\
        3: Compute \(\pi_i: \:\:\: p_{i} = P(sign(z_i)=1)=\frac{1}{2}(1+erf(-\frac{m_i}{\sqrt{2}v_i}))\)
        and \(1-p_i\) \\
        4: Sample \(g_i \sim Gumbel(0,1)\) \\
        5: Use Gumbel-Softmax to generate $\hat{h}_i$ (eq. \eqref{eqn:gumbel_Softmax})   \\
    \textbf{Return:} $h\in \mathbb{R}^{d_{out}}$ with $h_i=\hat{h}_{i,1} - \hat{h}_{i,-1}$   
\label{algo:algo1}    
\end{algorithm}

\subsection{Batch Normalization}
Batch normalization \citep{BN} is a common part of convolutional neural networks that is known to accelerate training and improve performance in many cases. Specifically for discrete networks, the authors in \citep{MPT} showed a high gain just by including batch normalization layers. In our model, before the sign activation function, we have the pre-activation distributions instead of deterministic values (see \ref{fig:1b}), and we cannot perform the classic batch normalization layer. \\



To address this issue, we propose a new batch normalization layer for our training process, which can be applied to distributions. The pre-activation \(z_{bcij}\) (\(b\) is the batch index and \(c\) is the channel index) of each convolution are approximated by Gaussian random variables and are distributed according to \(z_{bcij}\sim \mathcal{N}(\mu_{bcij},\sigma_{bcij}^2)\).
The mean of the normalized variable can be easily calculated as,
\begin{equation}
    \mu_{c}=E_{bij}[Z_{bcij}]=\frac{1}{B\cdot H \cdot W}\sum\limits_{b,i,j}\mu_{bcij}.
\end{equation}
Where $B,W,H$ are the batch size, the width, and the height of the image respectively. Using the Law of total variance, $Var(Y)=Var[E[Y|X]] + E[Var[Y|X]]$, we can also calculate the variance of the normalized variable: 
\begin{equation}
\sigma_{c}^2=\frac{1}{B\cdot H \cdot W}\sum\limits_{b,i,j}(\mu_{bcij}-\mu{_{c}})^2+
       \frac{1}{B\cdot H \cdot W}\sum\limits_{b,i,j}\sigma_{bcij}^2
\end{equation}    

Batch normalization usually also includes learnable affine parameters $\gamma$ and $\beta$, which we incorporate here as well. The pre-activation after batch normalization layer $z_{BN}$ is distributed according to

\begin{equation}
       z_{BN_{bcij}}\sim \mathcal{N}\left(\gamma_{c} \cdot \frac{\mu_{bcij}-\mu_{c}}{\sigma_{c}}+\beta_{c},\frac{\gamma^{2}_{c} \sigma_{bcij}^2}{\sigma_{c}^2}\right).
\end{equation}   


\subsection{Inference}
At inference or test time, we wish to apply our model with the actual discrete weights and activations and not the approximation used for training. To do that, we sample discrete weights based on the probabilities we obtained during training, and perform a standard inference using the sampled weights. This process can be repeated several times to identify the best-performing weights. However, it is worth noting that our trained distribution typically has low entropy, resulting in little variation in test accuracy between weight samples. As a result, only slight improvements in overall performance are observed.

\section{Implementation Details}
\subsection{Network Architecture}
We use the ResNet-18 \citep{resnet} and VGG \citep{vgg}, as is standard for experiments on discrete neural networks.
In all our experiments, similar to prior works, the first layer and last fully-connected layer remain in full precision. 


\subsection{Optimization Details}
\begin{figure}
  \begin{subfigure}{0.33\textwidth}
    \includegraphics[width=\linewidth]{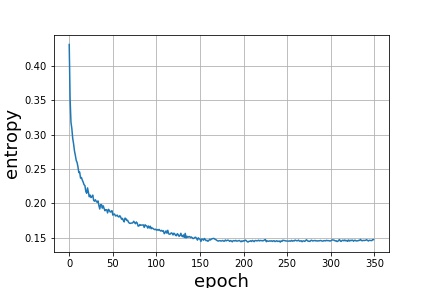}
    \caption{} \label{fig:entropy_a}
  \end{subfigure}%
  \hspace*{\fill}   
  \begin{subfigure}{0.33\textwidth}
    \includegraphics[width=\linewidth]{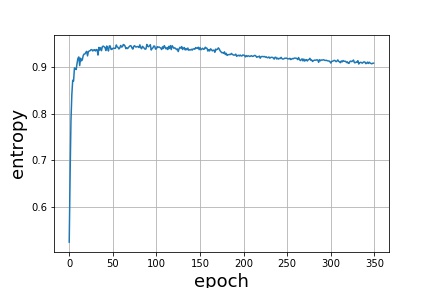}
    \caption{} \label{fig:entropy_b}
  \end{subfigure}%
  \hspace*{\fill}   
  \begin{subfigure}{0.33\textwidth}
    \includegraphics[width=\linewidth]{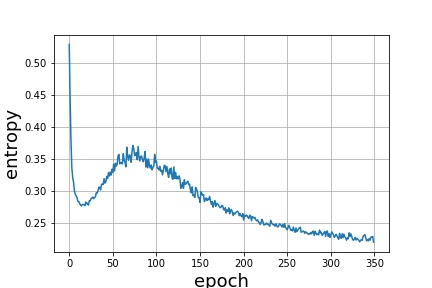}
    \caption{} \label{fig:entropy_c}
  \end{subfigure}
\caption{
(a) Activation entropy example from an intermediate layer of the network. The entropy is relatively low, meaning low randomness, but the model is not deterministic;
(b) Activation entropy example from the last binary layer, without learning rate decay;
(c) Example of activation entropy from the last binary layer after reducing the learning rate of the last FC layer.
\label{fig:entropy}
}
\end{figure}

We observed that incorporating multiple Monte-Carlo samples for each input element significantly enhances the overall results.
In every batch, we run our model on each datum multiple times, each with a different random sample resulting in the following gradient estimation
\begin{equation}
\label{eqn:sloss}
  \nabla_{\theta} L(\mathcal{\theta})\approx \sum_{i=1}^{N}\sum_{j=1}^{S}\nabla\ell(f(h_{i},\epsilon_j,\theta),y_{i}).
\end{equation}
\\
\paragraph{Weights Distributions Entropy.} Similar to \citep{Shayer01}, we added \(probability \:\: decay\) regularization. We used it as \(L_2\) regularization on the distribution parameters. It helps to increase the weights entropy and prevent many weights distributions converged to a deterministic value, which degrades the CLT approximation and leads to suboptimal results.
\paragraph{Activations Distributions Entropy.} In some cases, we observed that there is a need to reduce the entropy of the activations distributions of the last binary layer. We found that many of the activations probabilities $p_i$ in that layer converge to values around $0.5$, which means an undesirable extremely high amount of randomness. We found that giving a lower learning rate to the fully connected layer (after the last binary layer) helps to reduce the entropy. We show this phenomenon in Fig. \ref{fig:entropy}.
\paragraph{Initialization.} Similar to \citep{Shayer01} we use a pre-trained network to initialize the weight distribution parameters $\theta$. We found that using a network with \emph{Tanh} as an activation function instead of \emph{ReLU} results in a better initialization for $\theta$, as it is closer to our discrete sign activation. For further details on the initialization scheme, please refer to the supplementary material.



\section{Experiments}
In this section, we detail the conducted experiments. We use two benchmark datasets, CIFAR-10 and CIFAR-100 \citep{Krizhevsky01}, to verify the effectiveness of our proposed method and compare it with other state-of-the-art methods. Our code is available in the supplementary material.

\paragraph{CIFAR-10.} We compare our results with prior works on CIFAR-10, including IR-Net\citep{IRNet}, MPT \citep{MPT}, BNN-DL\citep{RAD}, DSQ\citep{DSQ},
BNN\citep{BNN} and XNOR-Net\citep{XNOR} using VGG-small and ResNet-18. We compared networks with 1-bit weights and 1-bit activations (1/1) as well as networks with 2-bits weights and 1-bit activations (2/1). The extra bit resulting in weights sparsity, which brings additional efficiency.
The performance comparison using the different methods is shown in Table \ref{res:cifar10} (we presented the best performance for each method over VGG-small and ResNet-18). The results show that our approach achieves better results than other existing methods. The entire experimental details are provided in the supplementary material. \\
\paragraph{CIFAR-100.} We also compared our model performance on CIFAR-100 with IR-Net\citep{IRNet}, MPT\citep{MPT}, and BNN-DL\citep{RAD}. 
The results are shown in Table \ref{res:cifar100}. Our approach achieves better results than other existing methods, 
with the exception of BNN-DL which on par. However, we note that the BNN-DL uses a slightly larger network. All the experimental details regarding our training process and the evaluation for MPT and IR-Net 
are included in the supplementary material.

\begin{table}
\begin{center}
\begin{tabular}{|c|c|c|c|c|}
\hline
Method & Topology & Bit-Width (W/A) & Top-1 Acc\\
\hline\hline
Full-Precision   & ResNet-18          & 32/32        & 95.01           \\
LR-net           & ResNet-18          & 2/32         & 94.88          \\
\hline
IR-Net           & ResNet-18          & 1/1          & 91.5            \\
BNN-DL           & ResNet-18          & 1/1          & 90.5            \\
DSQ              & VGG-small          & 1/1          & 91.72           \\
MPT              & VGG-small          & 2/1          & 91.9            \\
BNN              & VGG-small          & 1/1          & 89.9            \\
XNOR-Net         & VGG-small          & 1/1          & 89.8            \\
\textbf{LAR-net}  & \textbf{ResNet-18} & \textbf{2/1} & \textbf{92.34 $\pm$ 0.05} \\
\hline
\end{tabular}
\end{center}
\caption{Performance comparison with Binary SOTA methods on CIFAR-10.}
\label{res:cifar10}
\end{table}

\renewcommand*{\thefootnote}{\fnsymbol{footnote}}

\begin{table}
\begin{center}
\begin{tabular}{|c|c|c|c|c|}
\hline
Method & Topology & Bit-Width (W/A) & Top-1 Acc \\
\hline\hline
Full-Precision   & ResNet-18              & 32/32         & 75.61                      \\
LR-net           & ResNet-18              & 2/32          & 73.13                      \\
\hline
IR-Net           & ResNet-18              & 1/1           & 67.42 $\pm$ 0.06           \\
IR-Net           & VGG-small              & 1/1           & 67.8 $\pm$ 0.04            \\
BNN-DL           & ResNet-20\footnotemark & 1/1           & 68.17                      \\
MPT              & VGG-small              & 2/1           & 63.71 $\pm$ 0.16           \\
\textbf{LAR-net}  & \textbf{ResNet-18}     & \textbf{2/1}  & \textbf{68.35 $\pm$ 0.02} \\
\hline
\end{tabular}
\end{center}
\caption{Performance comparison with Binary SOTA methods on CIFAR-100.}
\label{res:cifar100}
\end{table}
\footnotetext{They used their variant of ResNet with 20 layers, we refer to this for convenience as resnet-20}

\section{Conclusions}
In this paper, we introduced a novel scheme for training a network with discrete weights and activations, for efficient inference of vision and learning applications. We demonstrate that using a probabilistic approach is not limited to networks with discrete weights, but can also be successfully applied to networks with discrete activations. Furthermore, we show how standard DNN layers, such as batch normalization can also be applied in this scenario. Finally, we evaluate our method on various image classification datasets obtaining state-of-the-art results. 



\bibliographystyle{plainnat}
\bibliography{egbib}








\newpage

\appendix

\section{LAR-Net Initialization}
\label{appendix:init}
In full-precision networks, it is common to use a random initializer for the weight initialization, e.g., Kaiming Normal~\citep{Kaiming01}. We are interested in initializing the distributions over discrete weights.
\citet{Shayer01} suggested using pretrained continuous deterministic weights for the distributions initialization. Let $\Tilde{W}$ be the normalized pretrained weights (by dividing the weights in each layer $t$ by the standard deviation of the weights \(\sigma^{(t)}\)). \\
Then, \(p(w_{ij}^{(t)}=0)\) is initialized by
\begin{equation}
    p(w_{ij}^{(t)}=0)=p_{max} - (p_{max} - p_{min}) \cdot \lvert \Tilde{w}_{ij}^{(t)} \rvert 
    \label{eqn:p0}
\end{equation}

Where $p_{max}$ and $p_{min}$ are hyperparameters (set to 0.05 and 0.95, respectively, in our experiments). Next, \(p(w_{ij}^{(t)}=1 \lvert w_{ij}^{(t)} \neq 0)\) is initialized by
\begin{equation}
    p(w_{ij}^{(t)}=1 \lvert w_{ij}^{(t)} \neq 0) = 0.5 \cdot ( 1 + \frac{\Tilde{w}_{ij}^{(t)}}{1-p(w_{ij}^{(t)}=0)} )    
    \label{eqn:p1}
\end{equation}
Then, all the values from equations \ref{eqn:p0},\ref{eqn:p1} are clipped in range $[ p_{min}, p_{max} ]$. 
We found that while this initialization works well also to train networks with discrete activations, we achieved better results when we used LR-Net (with only discrete weights) as a baseline. 
We first trained a network with discrete weights and then transfer the weights distributions as initialization to a network with discrete activations as well. 

\section{CIFAR-10 Experimental Details}
\subsection{Hyperparameters for LAR-Net}
\label{appendix:cifar10exp}
CIFAR-10~\citep{Krizhevsky01} is an image classification benchmark dataset. It consists of 50,000 training images and 10,000 test images distributed among ten different classes. Each image in the dataset is 32 × 32 pixels in RGB space, and it is preprocessed by subtracting its mean and dividing by its standard deviation. During training, we apply a padding of four pixels to each side of the image, and a random 32 × 32 crop is sampled from the padded image. Furthermore, we flip images horizontally at random during training. At test time, we evaluate the original 32 × 32 image without any padding or multiple cropping. The loss is minimized with Adam \citep{adam}. The weight decay parameter is set to \(1e-4\). 
We noticed that by executing multiple Monte-Carlo samples for every input element, we can improve the results. 
In each batch, our model is executed multiple times for each data, using a random sample, which results in eq. \ref{eqn:sloss}.
We use a batch size of 64 and run 2 times each batch ($S=2$). The initial learning rate is 0.01, and we use a cosine decay learning rate policy for training. The probability decay parameter is set to \(1e-12\), and the temperature of the Gumbel Soft-Max is 1.2 (fixed for all the training). We report the test error rate after 300 training epochs. Our discrete weights are \(\{-1,0,1\}\), which also brings additional savings. In our method, the sparsity level depends on the learned $\theta$ and is not pre-defined. We achieved a sparsity of \(44\%\pm1.1\%\).

\section{CIFAR-100 Experimental Details}

\subsection{Hyperparameters for LAR-Net}
\label{appendix:cifar100exp}
CIFAR-100~\citep{Krizhevsky01} is comprised of 60,000 32x32 color images that are grouped into 100 classes, with 600 images for each class. The dataset is divided into 50,000 training images and 10,000 testing images. The loss is minimized with Adam \citep{adam}. The weight decay parameter is set to \(1e-4\). We use a batch size of 32 and run 4 times every batch ($S=2$). The initial learning rate is 0.01, and we use a cosine decay learning rate policy for training. Similar to CIFAR-10, The probability decay parameter is set to \(1e-12\), and the temperature of the Gumbel Soft-Max is 1.2. We report the test error rate after 300 training epochs. We achieved a weight sparsity of \(43\%\pm1.0\%\).

\subsection{Hyperparameters for MPT and IR-Net}
\label{appendix:cifar100}

We searched over all the next hyper-parameters to find the best results on CIFAR-100. For IR-Net \citep{IRNet}, we used the code from \citep{IRnetGit}, and for MPT \citep{MPT}, we used the code from \citep{MPTGit}.
We tried to find the best results with ResNet-18 and VGG-small. We used Adam optimizer and SGD, and searched the learning rate (LR) in the range [0.1, 0.001]. We checked different batch sizes [64, 128, 256] and also different weight decay configurations. For MPT, we used the version in which the batch normalization parameters are also learned (which they called MPT+BN). They achieved better results with this version consistently. The weights were initialized using the Kaiming Normal \citep{Kaiming01}. 
Table \ref{res:CIFAR100_exp} presents the best configurations for the models we checked.

\begin{table}[htbp]
\begin{center}
\resizebox{\columnwidth}{!}{\begin{tabular}{cccccccc}
\hline
Method     & Model     & Optimizer    & LR   & Momentum & Weight Decay & Batch & Epochs \\
\hline
IR-Net     & ResNet-18 & SGD          & 0.035 &  0.9     &  1e-4        & 128   & 400     \\
IR-Net     & VGG-small & SGD          & 0.01  &  0.9     &  1e-4        & 128   & 400     \\
MPT + BN   & ResNet-18 & Adam         & 0.1   &          &  5e-4        & 128   & 400     \\
MPT + BN   & VGG-small & Adam         & 0.01  &          &  1e-4        & 128   & 400     \\
\hline
\end{tabular}}
\end{center}
\caption{Hyperparameter Configurations for CIFAR-10 Experiments}
\label{res:CIFAR100_exp}
\end{table}

\section{Ablation Study}
We investigate the performance of LAR-Net without a Batch Normalization layer. We separate our experiment into the different parts in the BN layer: the normalization (using $mean$ and $var$) and the affine transform (using learnable parameters $\gamma$ and $\beta$). 
Table \ref{res:bn_ablation} shows the results using the different settings. We use \textbf{no BatchNorm} to indicate no Batch Normalization layer at all, and \textbf{Affine Transform only} to indicate only affine transform using $\gamma$ and $\beta$, but without the normalization.

\begin{table}[htbp]
\begin{center}
\begin{tabular}{c|c|c|c}
\hline
Dataset     & Model     & BatchNorm & Top-1 Acc \\
\hline
           & ResNet-18 &  with BatchNorm         &  68.31  \\
CIFAR-100  & ResNet-18 &  no BatchNorm           &  60.71  \\
           & ResNet-18 &  Affine Transform only  &  61.82  \\
\hline
           & ResNet-18 &  with BatchNorm         &  92.34  \\
CIFAR-10   & ResNet-18 &  no BatchNorm           &  85.65  \\
           & ResNet-18 &  Affine Transform only  &  83.07  \\
\hline
\end{tabular}
\end{center}
\caption{Ablation Study for LAR-Net}
\label{res:bn_ablation}
\end{table}

\end{document}